\documentclass[conference]{IEEEtran}
\IEEEoverridecommandlockouts
\usepackage{cite}
\usepackage{amsmath,amssymb,amsfonts}
\usepackage{algorithmic}
\usepackage{graphicx}
\usepackage{textcomp}
\usepackage{xcolor}
\def\BibTeX{{\rm B\kern-.05em{\sc i\kern-.025em b}\kern-.08em
T\kern-.1667em\lower.7ex\hbox{E}\kern-.125emX}}

\makeatletter
\newcommand{\linebreakand}{%
  \end{@IEEEauthorhalign}
  \hfill\mbox{}\par
  \mbox{}\hfill\begin{@IEEEauthorhalign}
}
\makeatother

\begin{document}

\title{An Autotuning-based Optimization Framework for Mixed-kernel SVM Classifications in Smart Pixel Datasets and Heterojunction Transistors}


\author{\IEEEauthorblockN{Xingfu Wu} \IEEEauthorblockA{Mathematics and Computer Science Division \\ 
Argonne National Laboratory, Lemont, IL, USA \\ xingfu.wu@anl.gov}
\and
\IEEEauthorblockN{Tupendra Oli} \IEEEauthorblockA{High Energy Physics Division \\ 
Argonne National Laboratory, Lemont, IL, USA \\
toli@anl.gov}
\linebreakand 
\IEEEauthorblockN{Justin H. Qian} \IEEEauthorblockA{Department of Materials Science and Engineering \\ Northwestern University, Evanston, IL, USA\\ JustinQian2027@u.northwestern.edu}
\and
\IEEEauthorblockN{Valerie Taylor} \IEEEauthorblockA{Mathematics and Computer Science Division \\ Argonne National Laboratory, Lemont, IL, USA \\ vtaylor@anl.gov}
\linebreakand 
\IEEEauthorblockN{Mark C. Hersam} \IEEEauthorblockA{Department of Materials Science and Engineering \\ Northwestern University, Evanston, IL, USA\\ 
m-hersam@northwestern.edu}
\and
\IEEEauthorblockN{Vinod K. Sangwan} \IEEEauthorblockA{Department of Materials Science and Engineering \\ Northwestern University, Evanston, IL, USA\\ 
vinod.sangwan@northwestern.edu}
}

\maketitle

\begin{abstract}
Support Vector Machine (SVM) is a state-of-the-art classification method widely used in science and engineering due to its high accuracy, its ability to deal with high dimensional data, and its flexibility in modeling diverse sources of data. In this paper, we propose an autotuning-based optimization framework to quantify the ranges of hyperparameters in SVMs to identify their optimal choices, and apply the framework to two SVMs with the mixed-kernel between Sigmoid and Gaussian kernels for smart pixel datasets in high energy physics (HEP) and mixed-kernel heterojunction transistors (MKH). 
Our experimental results show that the optimal selection of hyperparameters in the SVMs and the kernels greatly varies for different applications and datasets, and choosing their optimal choices is critical for a high classification accuracy of the mixed kernel SVMs. Uninformed choices of hyperparameters C and coef0 in the mixed-kernel SVMs result in severely low accuracy, and the proposed framework effectively quantifies the proper ranges for the hyperparameters in the SVMs to identify their optimal choices to achieve the highest accuracy 94.6\% for the HEP application and the highest average accuracy 97.2\% with far less tuning time for the MKH application.
\end{abstract}


\section{Introduction}
Support Vector Machines (SVMs) are a set of supervised machine learning methods used for classification, regression, and outliner detection \cite{SVM, SVMS, scikit-learn}. The support vector method was first proposed for the case of pattern recognition \cite{Boser92}, and then generalized to regression \cite{Vapnik95, Vapnik96}. Due to its high accuracy, its ability to deal with high dimensional data, and its flexibility in modeling diverse sources of data, the SVM classifiers are widely used in pattern recognition, text classification, image classification, geographic information system, bioinformatics, biology, medical health, hand-written character recognition, face detection, classification of credit card fraud, and emerging science and engineering applications \cite{Bishop06, SVM, BW10, Noble06, CG20, MKH, NC23}. In this paper, we focus on autotuning two mixed-kernel SVM applications in processing smart pixel datasets \cite{SMD} in high energy physics (HEP) and mixed-kernel heterojunction transistors (MKH) \cite{MKH} for arrhythmia detection from electrocardiogram data \cite{MM01}.

SVMs are widely used in machine learning (ML) as it can handle both linear and nonlinear classifications. When most real-world data is not linearly separable, kernel functions are used as preprocessing methods to be applied to the training data to transform it to higher-dimensional space in order to make the data linearly separable. There are a number of different kernel types that can be applied to classify the data. Some popular kernel functions include polynomial kernel, Gaussian kernel (known as the radial basis function), and Sigmoid kernel. The Gaussian kernel usually outperforms the polynomial kernel in both accuracy and convergence time \cite{BW10}. Gaussian kernel has interpolation ability and is effective at identifying local properties, and Sigmoid kernel is better suited for identifying global characteristics but has a relatively weak interpolation ability \cite{MKH}. If these two kernels are combined linearly, SVMs with the mixed kernels between Gaussian and Sigmoid kernels not only take the advantages of both kernels but also often lead to the best classification accuracy. 

Using the mixed-kernel SVMs effectively requires a good understanding of how they work. When training a mixed kernel SVM model, a number of decisions is needed to be made: how to preprocess the data, what is the optimal mixed ratio between Gaussian and Sigmoid kernels? how to optimize hyperparameters in SVMs and kernels? Uninformed choices may result in severely reduced accuracy \cite{CL11, BW10}, however, there is little insight about the uninformed choices. Most approaches for SVM research and applications focused on tuning three main ratios --- mixed kernel ratio, Sigmoid ratio, and Gaussian ratio --- to find the best combination for the highest accuracy of the mixed-kernel SVMs, however, the selection of hyperparameters in the SVMs and the kernels may greatly vary for different applications and datasets, and choosing the optimal kernel is critical for a high classification accuracy of the mixed-kernel SVMs \cite{MKH}.  Exhaustively evaluating all hyperparameter combinations becomes very time-consuming. Therefore, autotuning for automatic exploration of the hyperparameter space is necessary. This is the main motivation of this work here. 

The regularization parameter C with the default value 1.0 is common to all SVMs, and the strength of the regularization for training data is inversely proportional to C \cite{SVMS}. The parameter coef0 with the default value 0.0 is significant in the kernel sigmoid \cite{SVMS}. Proper choices of C and coef0 are critical to the accuracy of the mixed-kernel SVMs. Nevertheless, little research has been reported in the SVM literature on both of these parameters. In \cite{BW10}, a smaller value of C (=10) allows to ignore points close to the boundary and increases the margin, but a large value of C (=100) decreases the margin. In \cite{CL11}, the performance impact of C with the values of 1 and 16 was discussed. Training with different values of the regularization parameter C was suggested as a useful trick in \cite{SS04}. v-SVR for different values of C (from 1, 2, 4, ..., to 4096) was discussed in \cite{SS00}. Different values of C were set for five different algorithms for the SVMs with the Sigmoid kernel in the experiments \cite{Platt99}. Overall, choices of C and coef0 were set based on researchers' experience, but little practical guidance has been presented about how to quantify the ranges of the parameters C and coef0 to set their proper choices. To address this challenge, in this paper we leverage our open source ML-based autotuning tool ytopt \cite{ytl} to conduct all autotuning experiments. We propose an autotuning-based optimization framework to quantify their ranges to identify their optimal choices, and apply the framework to the two mixed-kernel SVM applications to illustrate its effectiveness in achieving the highest accuracy and in identifying the best configuration for the SVMs.

In this paper, we make the following contributions:
\begin{itemize}
\item We propose an autotuning-based framework for optimization methodology of mixed-kernel SVM scientific simulations.
\item We apply the framework to optimize the hyperparameters of two scientific SVM simulations for the highest accuracy, and we compare their performance. We find that uninformed choices of hyperparameters C and coef0 in the mixed-kernel SVMs result in severely low accuracy.
\item We evaluate the effectiveness of the proposed framework to quantify the optimal ranges for five hyperparameters in SVMs and the kernels to avoid their uninformed choices and to identify their best configuration for the highest accuracy. 
\end{itemize}

The remainder of this paper is organized as follows. Section 2 describes the backgrounds about mixed-kernel SVM classifications. Section 3 proposes our autotuning-based optimization framework for the SVMs. Section 4 illustrates the optimization process to use the proposed framework to improve the accuracy of a HEP application and presents the experimental results. Section 5 discusses the accuracy of the MKH application and, based on what is learned from the HEP application, further improves its accuracy. Section 6 summarizes this paper and discusses some future work.

\section{Mixed-Kernel SVM Classifications}

SVM is a supervised machine learning method that classifies data by finding an optimal line or hyperplane that maximizes the distance between each class in a n-dimensional space. Since multiple hyperplanes can be found to differentiate classes, maximizing the margin between data points enables the method to find the best decision boundary between classes. This enables it to generalize to new data points and make accurate classification predictions. 

There are two approaches to calculating the maximum margin, or the maximum distance between classes, one is the hard-margin classification \cite{Boser92} and the other is soft-margin classification \cite{CV95}. If we use a hard-margin SVM, the data points will be perfectly separated outside of the support vectors. This is represented with the formula: $y_i(w^Tx_i +b) \ge 1 $ (i=1, 2,..., n), where the vector w is known as the weight vector, and b is called the bias for the given training vectors $x_i$ and $y_i$ is the label associated with $x_i$.  Our goal is to find w and b such that the geometric margin $1/{ \lVert w \rVert}$ is maximized or ${\lVert w \rVert}^2$ is minimized. This leads to the following constrained optimization problem:

\begin{align*} \label{eq:hm} 
&minimize_{w,b} 	\frac{1}{2} \lVert w \rVert^2   \\
& y_i(w^Tx_i +b) \ge 1  (i=1, 2,..., n) \tag{1} 
\end{align*} 

The constraints ensure that the maximum margin classifier classifies each data point correctly, which is possible because we assumed that the data is linearly separable. 

Soft-margin classification is more flexible, allowing the classifier to misclassify some data points through the use of slack variables $\xi_i$. To allow some misclassifications we replace the  inequality constraints in Equation \ref{eq:hm} with  $y_i(w^Tx_i +b) \ge 1-\xi_i  (i=1, 2,..., n) $, where $\xi_i$ allows a data point to be in the margin error ($ 0 \le \xi_i \le 1$) or to be misclassified ($\xi_i > 1$). Because a data point is misclassified if the value of its slack variable is greater than 1, $\sum^n_{i=1} \xi_i$ is a bound on the number of misclassified data points. The objective of maximizing the margin will be augmented with a term $C \sum^n_{i=1} \xi_i$ to penalize the misclassifications and margin errors. The optimization problem becomes the following:

\begin{align*} \label{eq:sm} 
&minimize_{w,b}  \frac{1}{2} \lVert w \rVert^2  + C \sum^n_{i=1} \xi_i \\
& y_i(w^Tx_i +b) \ge 1-\xi_i, \xi_i \ge 0  (i=1, 2,..., n) 	\tag{2} 
\end{align*} 

The constant regularization parameter C ($>0$) sets the relative importance of maximizing the margin and minimizing the amount of the slack. It adjusts the margin: a larger C value narrows the margin for the minimal misclassification while a smaller C value widens it, allowing for more misclassified data points. Based on the soft-margin classification, the support vector machines in scikit-learn \cite{SVMS} implemented support vector classifications (SVC) which are classes capable of performing binary and multi-class classification on a dataset. Its implementation is based on libsvm \cite{CL11}. The multiclass support is handled according to a one-vs-one scheme. 

SVMs are widely used in ML as it can handle both linear and nonlinear classifications, however, when the data is not linearly separable, kernel functions are used as preprocessing methods to be applied to the training data to transform it to higher-dimensional space in order to make the data linearly separable. There are a number of different kernel types that can be applied to classify data. Some popular kernel functions include polynomial kernel, Gaussian kernel, or Sigmoid kernel. One question is, which kernel should we use for processing our data? The Gaussian kernel usually outperforms the polynomial kernel in both accuracy and convergence time \cite{BW10}. Gaussian kernel has interpolation ability and is effective at identifying local properties, and Sigmoid kernel is better suited for identifying global characteristics but has a relatively weak interpolation ability \cite{MKH}. If these two kernels are combined linearly, the new mixed kernel function is
$K =(1-\alpha) K_1 + \alpha K_2$,
where $0 \le \alpha \le 1$ is the mixed kernel ratio, $K_1$ is the Sigmoid kernel function, $K_2$ is the Gaussian kernel function. If $\alpha = 1$, K is the Gaussian kernel function. If $\alpha = 0$, it is Sigmoid kernel function. 
In this way, the mixed kernel between Gaussian and Sigmoid kernels takes the advantages of both kernels. 

The class sklearn.svm.SVC in scikit-learn \cite{SVMS} is implemented based on libsvm \cite{CL11} and supports the mixed kernels between Gaussian and Sigmoid kernels. The function SVC has several hyperparameters. In this paper we focus on five main tunable hyperparameters mixed-ratio, sigmoid-ratio, gaussian-ratio, C, and coef0, and the others are used with their default settings. We define the mixed kernel function mixed\_kernel as 

{\it
(1 - mixed-ratio) * sigmoid(x, y, sigmoid-ratio, coef0) \\
+ mixed-ratio * gaussian(x, y, gaussian-ratio)
}

Where  sigmoid(x, y, sigmoid-ratio, coef0) is  {\it tanh(sigmoid-ratio * dot(x, y) + coef0)}, and gaussian(x, y, gaussian-ratio) is  
{\it exp( - gaussian-ratio * $\lVert x-y \rVert^2$)}. Notice that sigmoid-ratio is the parameter gamma for Sigmoid kernel, and gaussian-ratio is the parameter gamma for Gaussian kernel. 

We use the mixed kernel in the SVC as a python function like the parametrized one in our simulations as follows,

{\it
SVC(C, kernel=lambda x, y: mixed\_kernel(x, y, mixied-ratio, sigmoid-ratio, coef0, gaussian-ratio))  }     {\hfill  (3)} 

Where the parameter C with the default value 1.0 is regularization parameter and common to all SVM kernels. It trades off misclassification of training the data against simplicity of the decision surface. The strength of the regularization is inversely proportional to C. A low C makes the decision surface smooth, while a high C aims at classifying all training data correctly \cite{SVMS}. If the data has a lot of noisy observations, C should be decreased, which corresponds to more regularization. The parameter coef0 with the default value 0.0 is significant in the Sigmoid kernel. Choosing proper choices of C and coef0 is critical to the accuracy of the mixed-kernel SVMs.

\section{Autotuning-based Optimization Methodology for Mixed Kernel SVMs}
For mixed-kernel SVC, identifying the optimal kernel is critical for high classification accuracy. Since the optimal choice of hyperparameters of the SVC can significantly vary for different applications and datasets, exhaustively evaluating all hyperparameter combinations becomes very time-consuming. Therefore, autotuning for automatic exploration of the hyperparameter space is necessary. In this section, we use our ML-based autotuning tool ytopt \cite{ytl} to propose our autotuning-based optimization framework for the mixed kernel SVMs.

\subsection{Applying ytopt to autotune a SVM simulation}

ytopt \cite{ytl, WU23, WU22} is a ML-based autotuning software package that consists of sampling a small number of input parameter configurations, evaluating them, and progressively fitting a surrogate model over the input-output space until exhausting the user-defined wall clock time or the maximum number of evaluations. The package is built based on Bayesian Optimization that solves optimization problems.

\begin{figure}[ht]
\center
 \includegraphics[width=\linewidth]{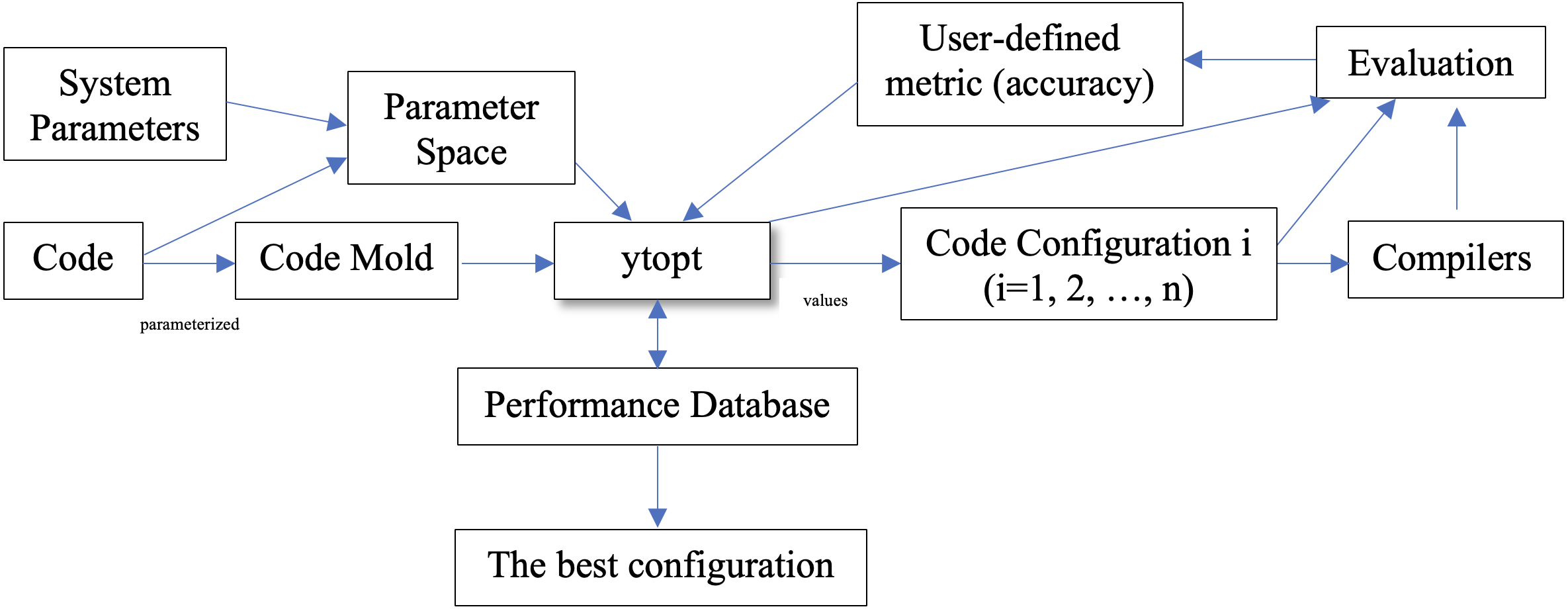}
 \caption{ytopt Autotuning Framework}
\label{fig:pf}
\end{figure}

Figure \ref{fig:pf} presents the autotuning framework ytopt.  We download ytopt and apply it to autotune a mixed-kernel SVM application. We define the accuracy of the SVM code as the performance metric. We analyze the SVM code to identify the five main tunable hyperparameters --- mixed-ratio, sigmoid-ratio, gaussian-ratio, C, and coef0 --- to define the parameter space using ConfigSpace \cite{CFS}. We use the tunable parameters to parameterize the SVM code as a code mold with the SVC function 3. 
ytopt starts with the user-defined parameter space, the code mold, and user-defined interface that specifies how to evaluate the code mold with a particular parameter configuration. 
The search method within ytopt uses Bayesian optimization, where a dynamically updated Random Forest surrogate model that learns the relationship between the configurations and the accuracy, is used to balance exploration and exploitation of the search space. In the exploration phase, the search evaluates parameter configurations that improve the quality of the surrogate model, and in the exploitation phase, the search evaluates parameter configurations that are closer to the previously found high-performing parameter configurations. The balance is achieved through the use of the lower confidence bound acquisition function that uses the surrogate models' predicted values of the unevaluated parameter configurations and the corresponding uncertainty values. 

\begin{figure}[ht]
  \centering
  \includegraphics[width=\linewidth]{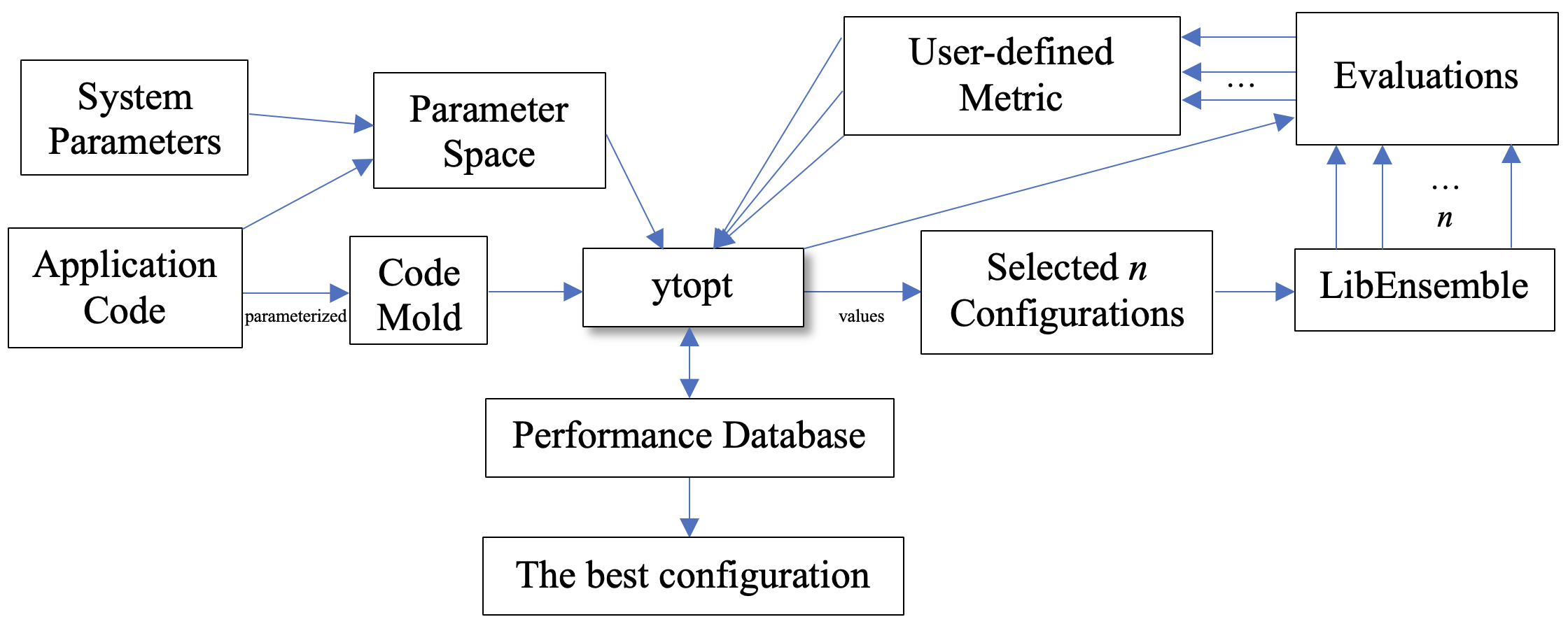}
  \caption{ytopt-libe Autotuning Framework}
  \label{fig:2}
\end{figure}

In our recent work \cite{WU22, WU23}, we developed and enhanced our autotuning framework ytopt to tune performance and energy for various scientific applications on large scale HPC systems. 
Figure \ref{fig:2} shows that we propose an asynchronously autotuning framework {\tt ytopt-libe} by integrating ytopt and libEnsemble \cite{WU24, ytl}. The framework {\tt ytopt-libe} comprises two asynchronous aspects:

\begin{itemize}
 \item The asynchronous aspect of the search allows the search to avoid waiting for all the evaluation results before proceeding to the next iteration. As soon as an evaluation is finished, the data is used to retrain the surrogate model, which is then used to bias the search toward the promising configurations.
\item The asynchronous aspect of the evaluation allows {\tt ytopt} to evaluate multiple selected parameter configurations in parallel by using the asynchronous and dynamic manager/worker scheme in {\tt libEnsemble}. All workers are independent; there is no direct communication among them.
When a worker finishes an evaluation, it sends its result back to  {\tt ytopt}. Then {\tt ytopt} updates the surrogate model and selects a new configuration for the worker to evaluate.
\end{itemize}

The advantages of this proposed autotuning framework {\tt ytopt-libe} are threefold:
\begin{itemize}
 \item Exploit the asynchronous and dynamic task management features provided by {\tt libEnsemble}
\item Accelerate the evaluation process of {\tt ytopt} to take advantage of massively parallel resources by overlapping multiple evaluations in parallel
\item Improve the accuracy of the random forest surrogate model by feeding more data to make the {\tt ytopt} search more efficient.
\end{itemize}

\subsection{Autotuning-based Framework}
As we focus on the five main tunable hyperparameters: mixed-ratio, sigmoid-ratio, gaussian-ratio, C, and coef0 in a mixed-kernel SVM code for autotuning using ytopt, based on our experience, we know the ranges for the mixed-ratio, sigmoid-ratio, and gaussian-ratio, but we do not have any ideas about the ranges for the parameters C and coef0, and we only know the default value of C is 1.0 and its values should be bigger than 0, and the default value of coef0 is 0.0 and its value may be positive or negative. 

\begin{figure}[ht]
  \centering
  \includegraphics[width=\linewidth]{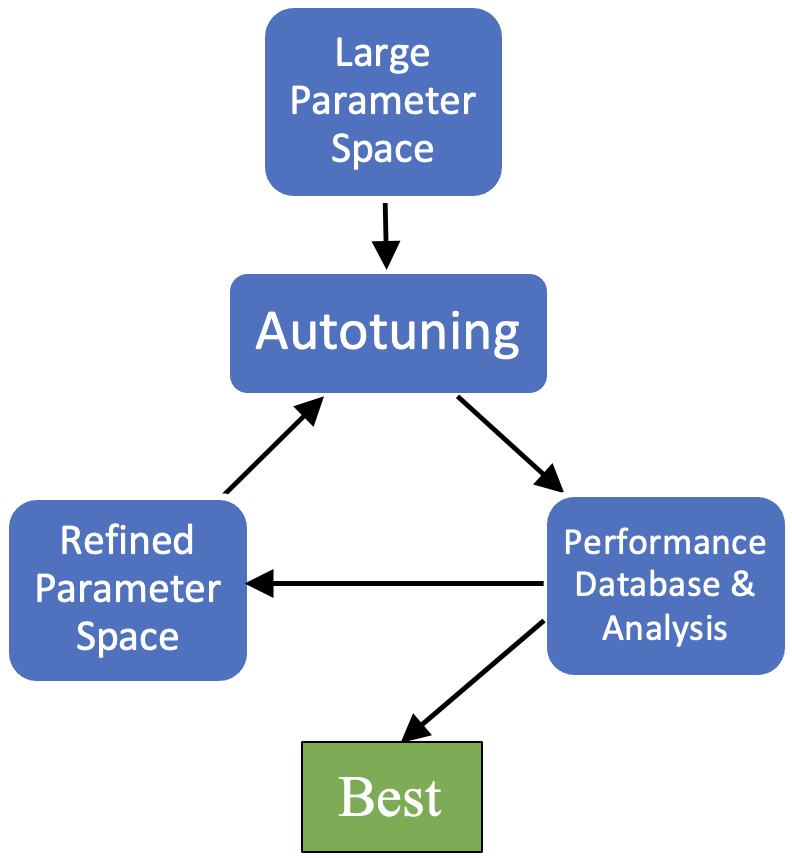}
  \caption{Autotuning-based Optimzation Framework}
  \label{fig:op}
\end{figure}

To quantify the ranges for the parameters C and coef0 in the SVM application in order to identify the optimal choices of C and coef0, we have to choose large ranges. For the parameter C, we set the large range [0.1, 100] with the quantization factor q (=0.01) because the value of C must be bigger than 0. For the parameter coef0, we set the large range [-15, 15] with the quantization factor q (=0.01). We define a large parameter space for the five hyperparameters. Then we use ytopt to autotune the SVM with the large parameter space, as shown in Figure \ref{fig:op}. The initial autotuning results (the configurations of five tunable parameters and their corresponding accuracy) for the SVM are stored in a performance database. After analyzing the accuracy with the corresponding configurations from the database, if the accuracy is improved over time, output the best accuracy with its corresponding configuration, and use the configuration as a reference to refine the parameter space to autotune the SVM again if needed. Otherwise, reduce the ranges for C and coef0 with the reduced quantization factor q (=0.001) to eliminate some values for C and coef0 which caused the worst accuracy, then use the refined parameter space to autotune the SVM again. Repeat this process if needed. Usually within a few iterations we not only quantify the proper ranges for C and coef0 but also identify the best configuration that will result in the highest accuracy for the SVM. 

In the rest of this paper, we apply the proposed autotuning-based optimization framework to two mixed-kernel SVM applications to evaluate its effectiveness. We conduct all our experiments on a computer server with Apple M1 Max chip with 10-core CPU and 32-core GPU and 64 GB of memory and 4 TB of hard drive.

\section{Smart Pixel Datasets}

We utilized datasets from futuristic pixel detectors in HEP experiments\cite{SMD}, which is a simulated dataset of silicon pixel clusters produced by charge particle (pions). This dataset comprises the charge deposited in each pixel per time slice upon pion interaction with the silicon pixel detectors, along with the truth information of each interacting pion. We implemented a SVM model with a mixed-kernel approach to effectively distinguish between low and high transverse momentum tracks, enabling the real-time filtration of undesirable low transverse momentum tracks. 

In this work, we use SVC from scikit-learn \cite{SVMS} to implement the mixed kernel SVM simulation to process the Smart Pixel datasets \cite{SMD}. We call it the HEP application. We identify the following five tunable hyperparameters: mixed-ratio, sigmoid-ratio, gaussian-ratio, C, and coef0, and define their hyperparameter space, then we use the proposed autotuning-based optimization framework to optimize the HEP application to identify the best combination of these parameters which results in the best accuracy.

For this mixed kernel SVM simulation, we apply ytopt to auotuning the HEP application with 128 evaluations. For the baseline accuracy, we use a default setting for the three ratios with C = 1.0, and coef0 = 0.0 to run the mixed kernel SVM to obtain its accuracy 83.4\%. The default setting was the best setting resulted from grid search. In our previous work \cite{WU23a}, ytopt outperformed grid search in accuracy and the tuning time.

\begin{figure}[ht]
\center
 \includegraphics[width=\linewidth]{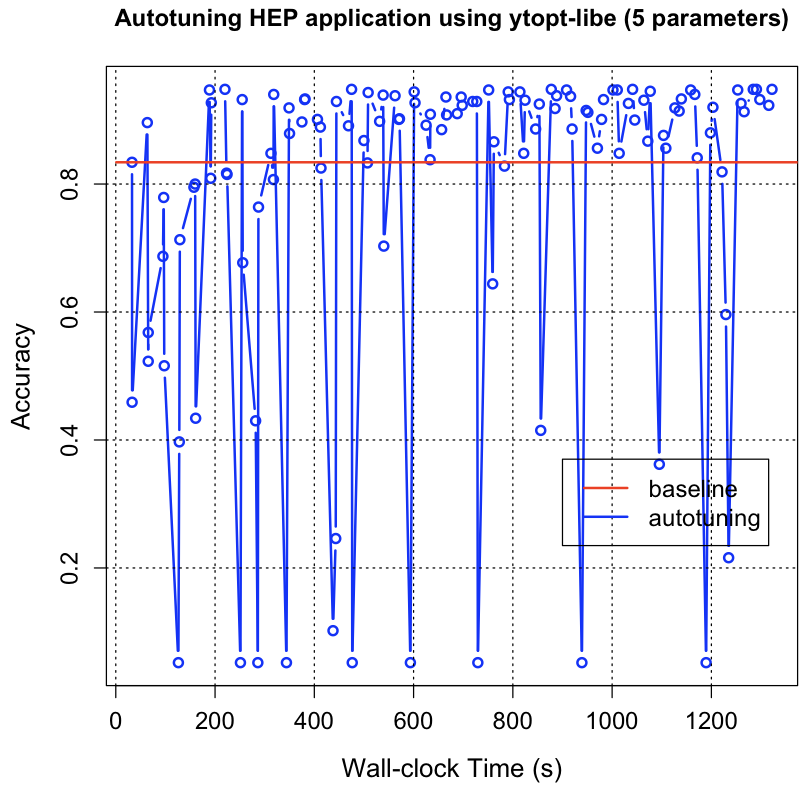}
 \caption{Autotuning the HEP application with 5 paramenters}
\label{fig:h5}
\end{figure}

Figure \ref{fig:h5} shows the autotuning process over time using five hyperparameters: mixed-ratio, sigmoid-ratio, and gaussian-ratio plus the parameters C and coef0, where the red line stands for the baseline accuracy, and each blue circle stands for the accuracy for the evaluation of a selected configuration. We set the range [0.1, 100] for the parameter C and the range [-15, 15] for the parameter coef0 with the quantization factor q=0.01. We observe some abnormal cases for which the accuracy was very low (5.2\%). This motivates us to investigate what really happened here. 

\begin{figure}[ht]
\center
 \includegraphics[width=\linewidth]{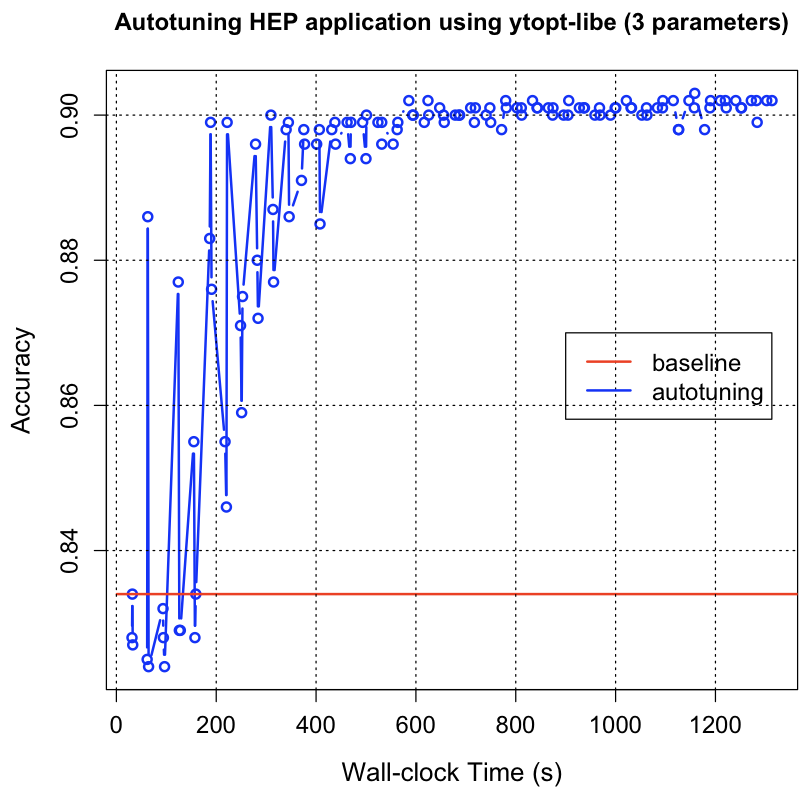}
 \caption{Autotuning HEP application with 3 paramenters}
\label{fig:h3}
\end{figure}

Then we use the parameter space for three hyperparameters: mixed-ratio, sigmoid-ratio, and gaussian-ratio with the default C = 1.0, and coef0 = 0.0 to autotune the mixed-kernel SVM. Figure \ref{fig:h3} shows the autotuning process over time. We observe that ytopt leads to the high performing region over time and achieved the best accuracy of 90.3\%.
Figure \ref{fig:h3} indicates that the three hyperparameters mixed-ratio, sigmoid-ratio, and gaussian-ratio are not the factor which resulted in the worst accuracy shown in Figure \ref{fig:h5}. This indicates either C or coef0 caused the worst accuracy.

\begin{figure}[ht]
\center
 \includegraphics[width=\linewidth]{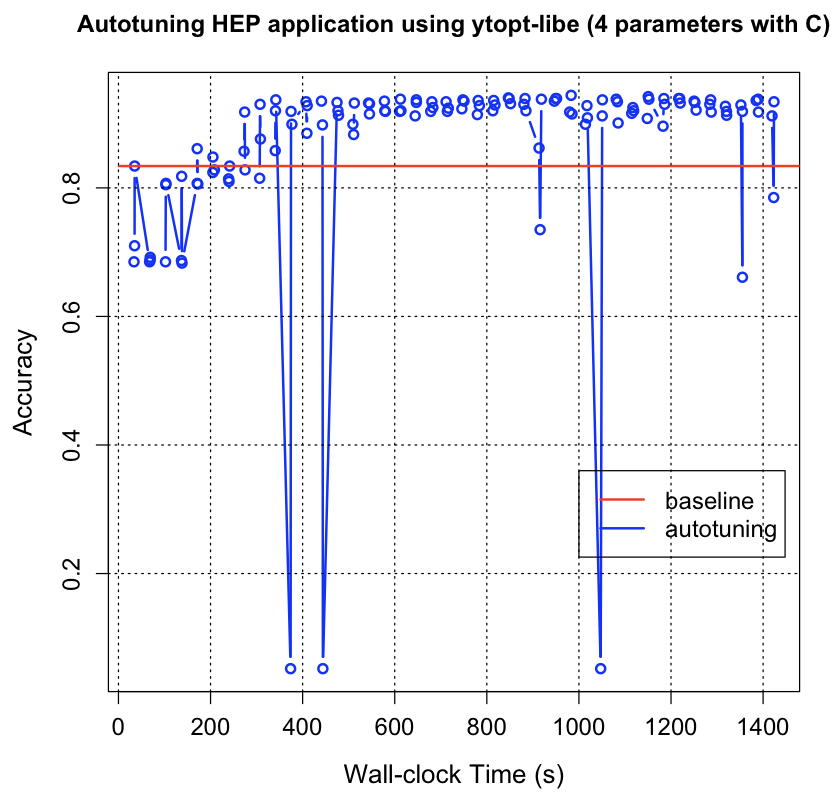}
 \caption{Autotuning HEP application with 4 paramenters with C and its range [0.1, 100]}
\label{fig:h4c0}
\end{figure}

First, based on the three parameters in Figure \ref{fig:h3}, we add the one more parameter C with the range [0.1, 100] with the quantization factor q=0.01 to tune the application again, as shown in Figure \ref{fig:h4c0}. 

\begin{figure}[ht]
\center
 \includegraphics[width=\linewidth]{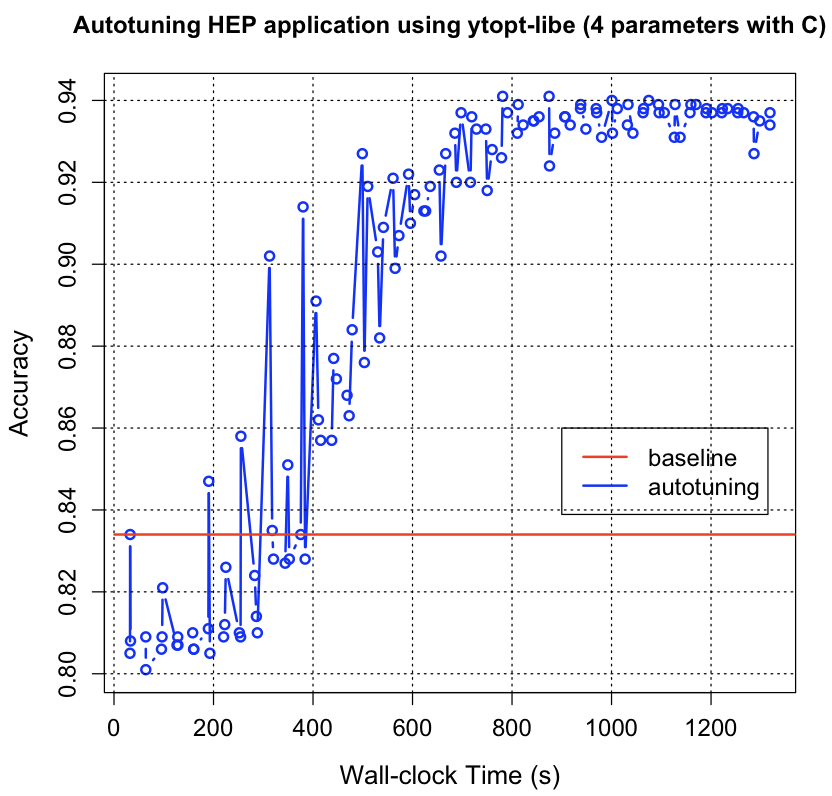}
 \caption{Autotuning HEP application with 4 paramenters with C and its refined range [0.37, 10]}
\label{fig:h4c}
\end{figure}

Based on the performance results for this autotuning, we find that when C$\leq$0.36, it caused the worst accuracy for some configurations; and when C $>$ 10, it did not result in the best accuracy. We reduce the range for the parameter C to [0.37, 10] with the reduced quantization factor q=0.001 to eliminate these worst configurations, then use ytopt to autotune the application again, as shown in Figure \ref{fig:h4c}. It reaches to the high performing region over time and results in the best accuracy of 94.1\%.

\begin{figure}[ht]
\center
 \includegraphics[width=\linewidth]{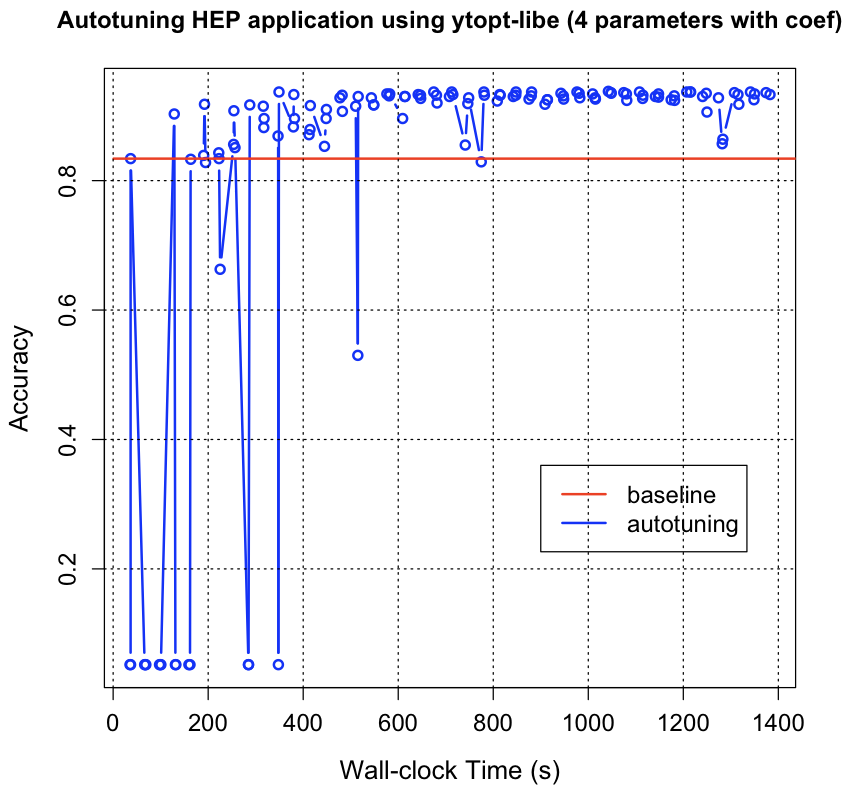}
 \caption{Autotuning HEP application with 4 paramenters with coef0 and its range [-15,15]}
\label{fig:h4e0}
\end{figure}

Second, based on the three parameters in Figure \ref{fig:h3}, we add the one more parameter coef0 with the range [-15, 15] with the quantization factor q=0.01 to tune the application again, as shown in Figure \ref{fig:h4e0}. 

\begin{figure}[ht]
\center
 \includegraphics[width=\linewidth]{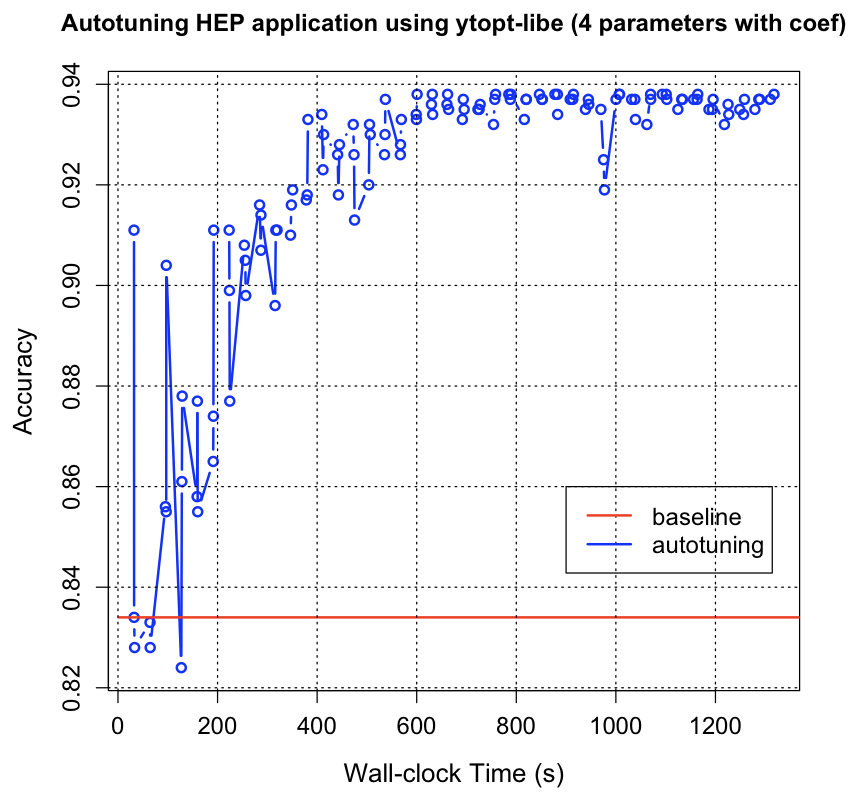}
 \caption{Autotuning HEP application with 4 paramenters with coef0 and its refined range [-1,1]}
\label{fig:h4e}
\end{figure}

Based on the performance results for this autotuning, we find that when coef0 $<$  -1 or coef0 $>$ 1, it caused the worst accuracy for some configurations. We reduce the range for the parameter coef0 to [-1, 1] with the reduced quantization factor q=0.001 to eliminate these worst configurations, then use ytopt to autotune the application again shown in Figure \ref{fig:h4e}. This time it reaches to the high performing region over time and results in the best accuracy of 93.8\%.

\begin{figure}[ht]
\center
 \includegraphics[width=\linewidth]{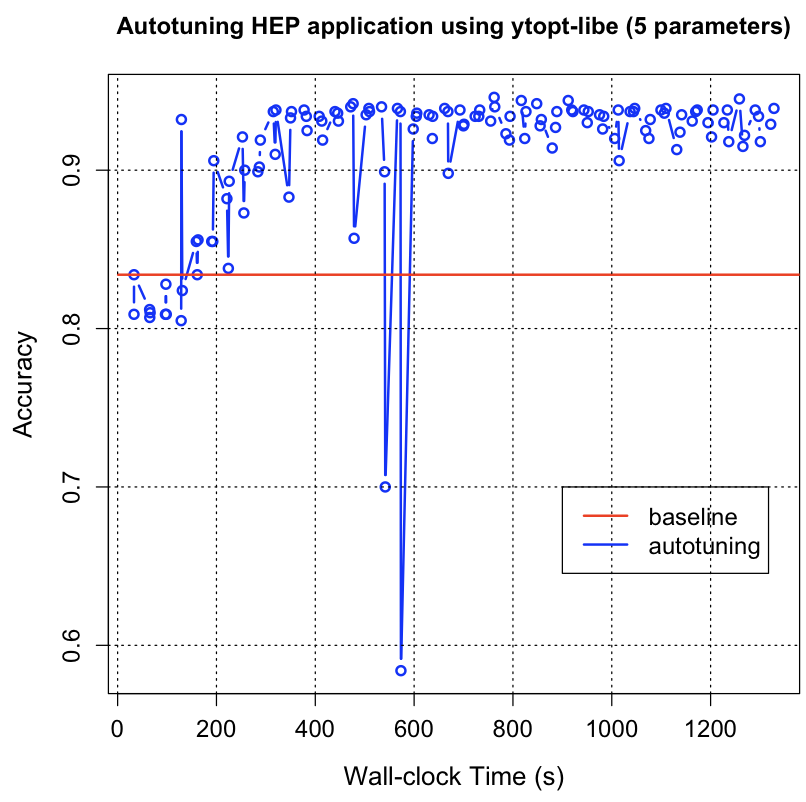}
 \caption{Autotuning HEP application with 5 paramenters with refined ranges}
\label{fig:h5r}
\end{figure}

From Figures \ref{fig:h4c} and \ref{fig:h4e}, we observe that the parameter C impacts the accuracy of the mixed-kernel SVMs more than the parameter coef0 does.
Next, we combine these refined ranges together to use the five hyperparameters to autotune the HEP application again shown in Figure \ref{fig:h5r}. This shows that the proposed framework leads the search to the high performing region to identify the best configuration over time. We achieve the best accuracy of 94.6\% with the configuration {\it \{mixed-ratio: 0.00036, sigmoid-ratio: 0.00017, gaussian-ratio:1.26, coef0: 0.8500000000000001, C: 0.37\}} for the application. The mixed-ratio: 0.00036 means that the mixed kernel function weights Sigmoid kernel more than Gaussian kernel. The regularization parameter C: 0.37 indicates that the smart pixel dataset is more noisy and needs more regularization.

\section{Mixed-kernel Heterojunction Transistors}

In SVM hardware implementations in \cite{MKH}, the authors reported dual-gated mixed-kernel heterojunction transistors using monolayer molybdenum disulfide as the n-type material and solution-processed semiconducting carbon nanotubes as the p-type material. They showed that  precise control over the electric field screening in MKH transistors enables the generation of a complete set of fine-grained Gaussian, sigmoid, and mixed-kernel functions using only a single device. They applied Bayesian optimization to enhance the mixed-kernel SVM classification accuracy of arrhythmia detection from electrocardiogram  data using the MKH transistors. The arrhythmia records as the input dataset to the mixed-kernel SVM system are from the publicly available MIT-BIH arrhythmia database \cite{MM01}.  They used SVC from scikit-learn to implement the mixed-kernel SVM simulation to detect six different types of arrhythmia \cite{MKH}: class (0) normal beat, class (1) atrial premature beat, class (2) premature ventricular contraction, class (3) paced beat, class (4) left bundle branch block beat, and class (5) right bundle branch block beat. They used the average arrhythmia detection accuracy with all six arrhythmia types as the main performance metric (accuracy) to identify which configuration results in the best accuracy.

\begin{figure}[ht]
\center
 \includegraphics[width=\linewidth]{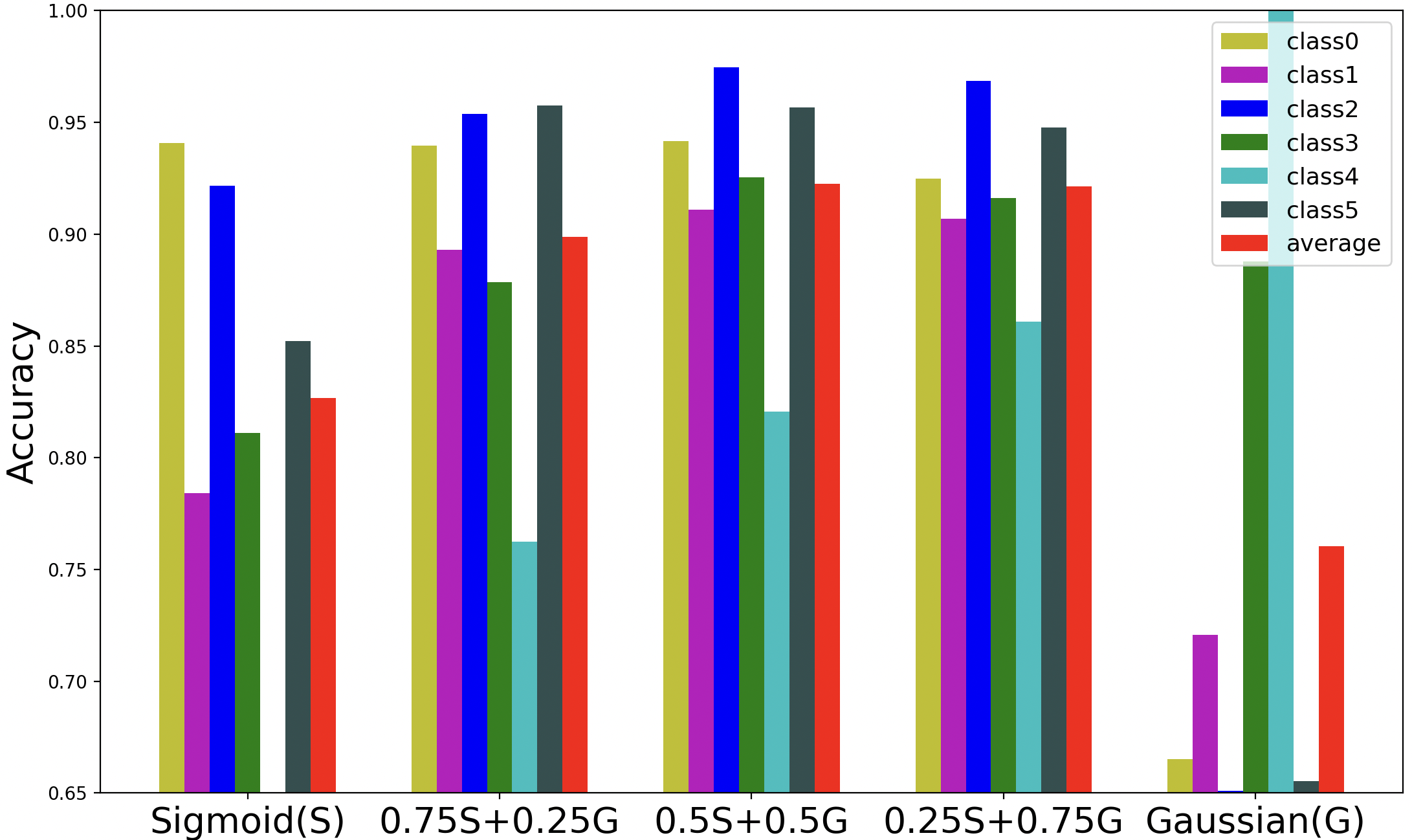}
 \caption{Classification accuracy for different mixed kernels}
\label{fig:mh1}
\end{figure}

We download the mixed-kernel heterojunction SVM simulation code from the GitHub repo \cite{YAN23} used in the MKH application. Figure \ref{fig:mh1} shows the classification accuracy for different mixed kernels with the mixed kernel ratios: 0.0, 0.25, 0.5, 0.75, and 1.0. We observe that the best average accuracy for the SVM with the mixed-ratio 0.5 is 92.3\%, which is used as the baseline. Notice that the SVM with the Sigmoid kernel outperformed that with the Gaussian kernel in the average accuracy (red bar). In \cite{MKH}, they focused mainly on a single parameter mixed-ratio and used Bayesian Optimization to tune the three parameters --- mixed-ratio, sigmoid-ratio, and gaussian-ratio --- to achieve the arrhythmia detection accuracies approaching 95\% with more than 2 hours tuning time for 25 evaluations. This motivates us to use the proposed framework to optimize the MKH application to compare the results.

Based on the mixed kernel SVM simulation code for the MKH application \cite{YAN23}, we apply ytopt to tune the application and compare the results with that from \cite{MKH}.
First, we focus on autotuning using one and three parameters with 128 evaluations to compare the results.

\begin{figure}[ht]
\center
 \includegraphics[width=\linewidth]{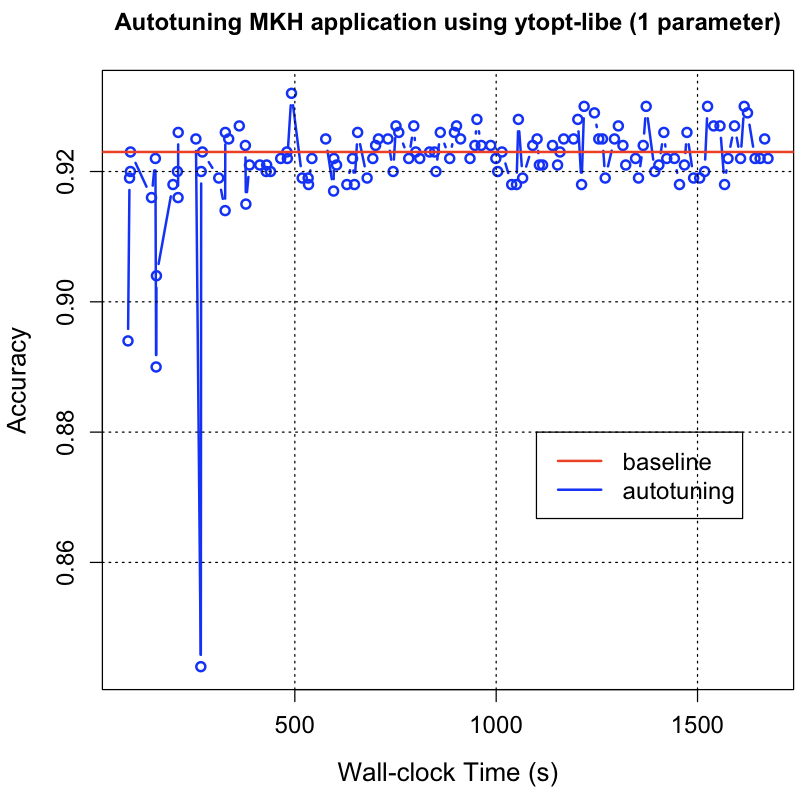}
 \caption{Autotuning MKH application with 1 paramenter}
\label{fig:m1}
\end{figure}

When we use a single hyperparameter mixed-ratio to autotune the application in Figure \ref{fig:m1}, we can only achieve the best accuracy of 93.2\%. We observe that no matter how many evaluations we use, the accuracy for the selected configurations is just little bit above the baseline accuracy. This indicates that just tuning the mixed-ratio did not lead to the best accuracy.

\begin{figure}[ht]
\center
 \includegraphics[width=\linewidth]{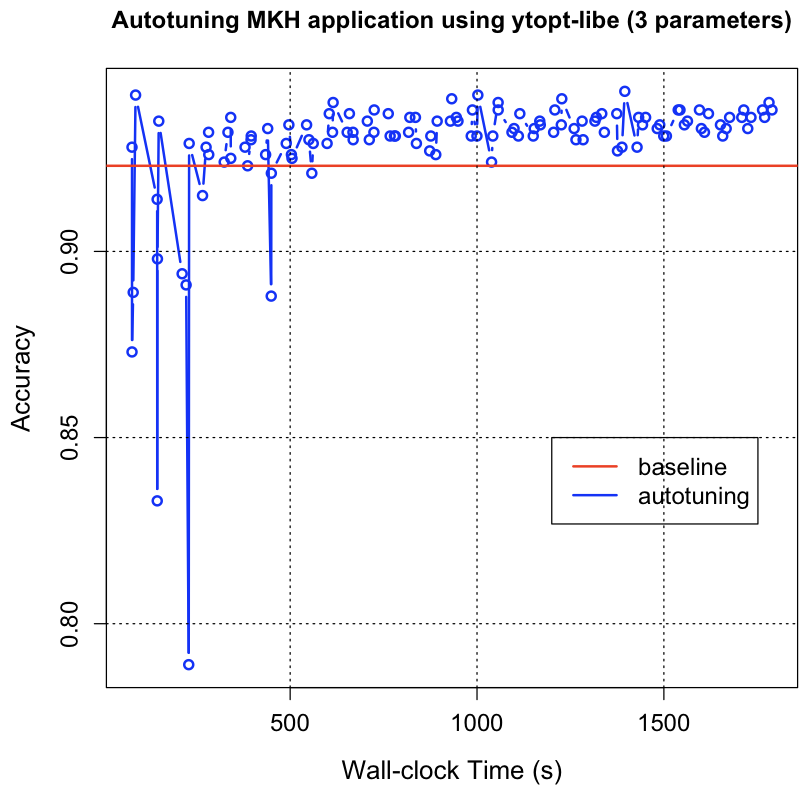}
 \caption{Autotuning MKH application with 3 paramenter}
\label{fig:m3}
\end{figure}

When we use three parameters: mixed-ratio, sigmoid-ratio, and gaussian-ratio to autotune the application in Figure \ref{fig:m3}, we can achieve the best accuracy of 94.3\%. We observe that the accuracy for the selected configuration is approaching 95\%. This trend is similar to that in \cite{MKH}, however, it took just less than a half hour to finish 128 evaluations in Figure \ref{fig:m3} comparing with more than 2 hours for only 25 evaluations in \cite{YAN23}. 

\begin{figure}[ht]
\center
 \includegraphics[width=\linewidth]{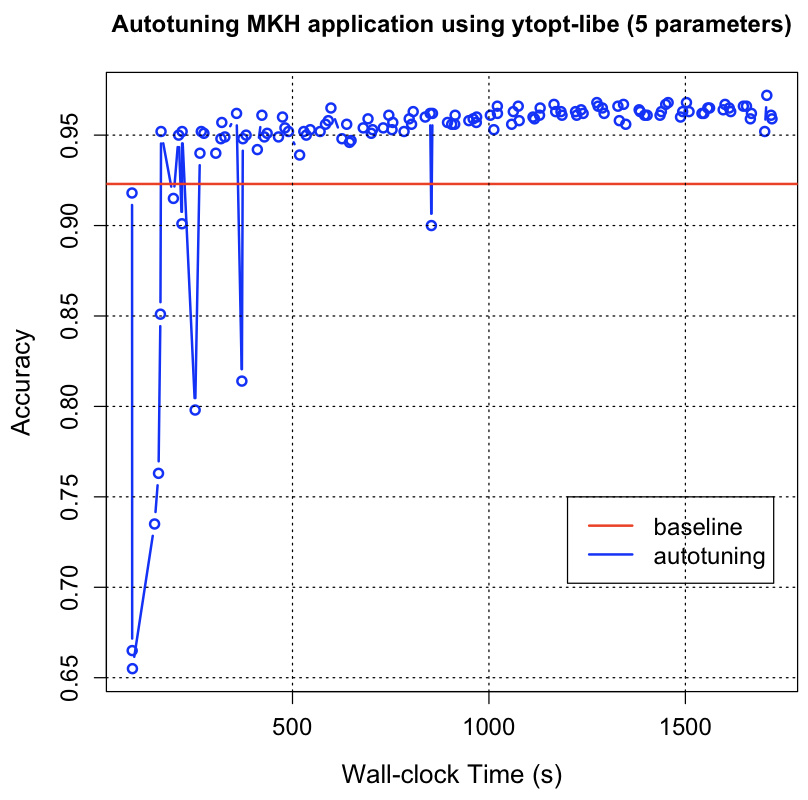}
 \caption{Autotuning MKH application with 5 paramenter}
\label{fig:m5}
\end{figure}

Based on what we learned from optimizing the HEP application, we further improve the accuracy of the MKH application by adding two more parameters C and coef0 into the mixed kernel SVM simulation code. As discussed in \cite{MKH}, the MKH transistors were amenable to personalized kernels that provided arrhythmia detection accuracies approaching 95\% for diverse patient profiles. When we add two more hyperparameters C with the range [0.37, 10] and coef0 with the range [-1, 1] to the parameter space to autotune the application, as shown in Figure \ref{fig:m5}, we achieve above 95\% arrhythmia detection accuracies over time. The best accuracy achieved in 128 evaluations is 97.2\% with the configuration {\it \{mixed-ratio:0.2085, sigmoid-ratio: 0.0001353314616106, gaussian-ratio: 0.8574993038146548,  coef0:-0.62, C:9.64\}}. The mixed-ratio: 0.2085 means that the mixed kernel function weights Sigmoid kernel more than Gaussian kernel. The regularization parameter C: 9.64 indicates that the data is less noisy and needs less regularization.This is really a big improvement in the average accuracy for the MKH application. 

Compared with the best configuration for the HEP application, we observe that the best selection of hyperparameters in SVMs can greatly vary for different applications and datasets, and choosing their optimal choices is critical for a high classification accuracy of the mixed kernel SVMs.  For the regularization parameter C, the best configuration for the HEP application chose C=0.37 which means more regularization because of the noisy data, but the best configuration for the MKH application had C=9.64 which indicates less regularization. It is interesting to see that, for the parameter coef0, its value is positive (0.85) for the HEP application but negative (-0.62)  for the MKH application.

\section{Conclusions}

Based on the two SVMs with the mixed-kernel between Sigmoid and Gaussian kernels, namely the HEP application and the MKH application, we can see that the optimal selection of hyperparameters in the SVMs and the kernels greatly varied for different applications and datasets, and choosing their optimal choices was critical for a high classification accuracy of the mixed-kernel SVMs. In the SVMs, the hyperparameters C and coef0 and their performance impact were less investigated in the literature, and their settings were just based on experience. Our experimental results show uninformed choices of them resulted in severely low accuracy (5.2\%). To address this challenge, in this paper we proposed the autotuning-based optimization framework to quantify their ranges to identify their optimal choices, and applied the framework to the two mixed-kernel SVM applications to illustrate its effectiveness. Our experimental results show that the proposed framework effectively quantified the proper ranges for the hyperparameters in the SVMs to identify their optimal choices to achieve the highest accuracy 94.6\% for the HEP application and the highest average accuracy 97.2\% with far less tuning time for the MKH application compared with less than 95\% average accuracy presented in \cite{MKH}.  The implementation codes for both applications are available on our GitHub repo \cite{YSVM}. The proposed autotuning-based optimization framework is flexible about which autotuning tool is used, in other words, other autotuning tools such as OpenTuner \cite{AK14}, GPTune \cite{LS21}, Bliss \cite{RT21} should be plug and play in the optimization framework. 

This work paves the way for us to integrate the whole stacks from the HEP application to the devices in our DoE ASCR Microelectronics project Threadwork \cite{TWORK}.
For the future work, we plan to use low-power mixed-kernel heterojunction transistors to process the smart pixel datasets so that we not only achieve high classification accuracy but also save the power consumption compared with current CMOS transistors for the detector data processing. 

\section*{Acknowledgments}

This work was supported in part by DOE ASCR Threadwork project, in part by DOE ASCR RAPIDS2 and OASIS, and in part by the National Science Foundation Materials Research Science and Engineering Center at Northwestern University under Award Number DMR-2308691. We acknowledge the Argonne Leading Computing Facilities (ALCF) for use of the GPU cluster Polaris under ALCF project EE-ECP. This material is based upon work supported by the U.S. Department of Energy, Office of Science, under contract number DE-AC02-06CH11357. 

\bibliographystyle{IEEEtran}
\bibliography{main}

\begin{thebibliography}{10}
\providecommand{\url}[1]{#1}
\csname url@samestyle\endcsname
\providecommand{\newblock}{\relax}
\providecommand{\bibinfo}[2]{#2}
\providecommand{\BIBentrySTDinterwordspacing}{\spaceskip=0pt\relax}
\providecommand{\BIBentryALTinterwordstretchfactor}{4}
\providecommand{\BIBentryALTinterwordspacing}{\spaceskip=\fontdimen2\font plus
\BIBentryALTinterwordstretchfactor\fontdimen3\font minus
  \fontdimen4\font\relax}
\providecommand{\BIBforeignlanguage}[2]{{%
\expandafter\ifx\csname l@#1\endcsname\relax
\typeout{** WARNING: IEEEtran.bst: No hyphenation pattern has been}%
\typeout{** loaded for the language `#1'. Using the pattern for}%
\typeout{** the default language instead.}%
\else
\language=\csname l@#1\endcsname
\fi
#2}}
\providecommand{\BIBdecl}{\relax}
\BIBdecl

\bibitem{SVM}
IBM, ``What are support vector machines (svms)?''
  https://www.ibm.com/topics/support-vector-machine, 2023.

\bibitem{SVMS}
scikit-learn 1.4.2, ``Support vector machines,''
  https://scikit-learn.org/stable/modules/svm.html, 2023.

\bibitem{scikit-learn}
F.~Pedregosa, G.~Varoquaux, A.~Gramfort, V.~Michel, B.~Thirion, O.~Grisel,
  M.~Blondel, P.~Prettenhofer, R.~Weiss, V.~Dubourg, J.~Vanderplas, A.~Passos,
  D.~Cournapeau, M.~Brucher, M.~Perrot, and E.~Duchesnay, ``Scikit-learn:
  Machine learning in {P}ython,'' \emph{Journal of Machine Learning Research},
  vol.~12, pp. 2825--2830, 2011.

\bibitem{Boser92}
B.~E. Boser, I.~M. Guyon, and V.~N. Vapnik, ``A training algorithm for optimal
  margin classifiers,'' in \emph{Proceedings of the 5th Annual ACM Workshop on
  Computational Learning Theory}, 1992.

\bibitem{Vapnik95}
V.~N. Vapnik, \emph{The nature of statistical learning theory}.\hskip 1em plus
  0.5em minus 0.4em\relax Springer-Verlag, New York, 1995.

\bibitem{Vapnik96}
\BIBentryALTinterwordspacing
V.~N. Vapnik, S.~E. Golowich, and A.~Smola, ``Support vector method for
  function approximation, regression estimation and signal processing,'' in
  \emph{Neural Information Processing Systems}, 1996. [Online]. Available:
  \url{https://api.semanticscholar.org/CorpusID:19196574}
\BIBentrySTDinterwordspacing

\bibitem{Bishop06}
C.~M. Bishop, \emph{Pattern Recognition and Machine Learning}.\hskip 1em plus
  0.5em minus 0.4em\relax Springer-Verlag, 2006.

\bibitem{BW10}
A.~Ben-Hur and J.~Weston, \emph{A User's Guide to Support Vector
  Machines}.\hskip 1em plus 0.5em minus 0.4em\relax Totowa, NJ: Humana Press,
  2010, pp. 223--239.

\bibitem{Noble06}
W.~S. Noble, ``A biologist’s introduction to support vector machines,''
  Department of Genome Sciences Department of Computer Science and Engineering
  University of Washington, 2006.

\bibitem{CG20}
\BIBentryALTinterwordspacing
J.~Cervantes, F.~Garcia-Lamont, L.~Rodríguez-Mazahua, and A.~Lopez, ``A
  comprehensive survey on support vector machine classification: Applications,
  challenges and trends,'' \emph{Neurocomputing}, vol. 408, pp. 189--215, 2020.
  [Online]. Available:
  \url{https://www.sciencedirect.com/science/article/pii/S0925231220307153}
\BIBentrySTDinterwordspacing

\bibitem{MKH}
\BIBentryALTinterwordspacing
X.~Yan, J.~H. Qian, J.~Ma, A.~Zhang, S.~E. Liu, M.~P. Bland, K.~J. Liu,
  X.~Wang, V.~K. Sangwan, H.~Wang, and M.~C. Hersam, ``Reconfigurable
  mixed-kernel heterojunction transistors for personalized support vector
  machine classification,'' \emph{Nature Electronics}, vol. Volume 6, 2023.
  [Online]. Available: \url{https://doi.org/10.1038/s41928-023-01042-7}
\BIBentrySTDinterwordspacing

\bibitem{NC23}
K.~Nagar and M.~Chawla, ``A survey on various approaches for support vector
  machine based engineering applications,'' \emph{International Journal of
  Emerging Science and Engineering}, vol.~11, 2023.

\bibitem{SMD}
M.~Swartz and J.~Dickinson, ``Smart pixel dataset (version 1) [data set],''
  https://doi.org/10.5281/zenodo.7331128, 2022.

\bibitem{MM01}
G.~B. Moody and R.~G. Mark, ``The impact of the mit-bih arrhythmia database:
  History, lessons learned, and its influence on current and future
  databases,'' \emph{IEEE ENGINEERING IN MEDICINE AND BIOLOGY}, vol. May/June,
  pp. 45–--50, 2001.

\bibitem{CL11}
C.~Chang and C.-J. Lin, ``Libsvm: A library for support vector machines,''
  \emph{ACM Transactions on Intelligent Systems and Technology}, vol. Volume 2,
  Issue 3, pp. 1–--27, 2011.

\bibitem{SS04}
A.~J. SMOLA and B.~SCHOLKOPF, ``A tutorial on support vector regression,''
  \emph{Statistics and Computing}, vol.~14, pp. 199–--222, 2004.

\bibitem{SS00}
B.~SCHOLKOPF, A.~J. SMOLA, R.~C. Williamson, and P.~L. Bartlett, ``New support
  vector algorithms,'' \emph{Neural Computation}, vol.~12, pp. 1207–--1245,
  2000.

\bibitem{Platt99}
J.~C. Platt, \emph{Probabilistic Outputs for Support Vector Machines and
  Comparisons to Regularized Likelihood Methods}.\hskip 1em plus 0.5em minus
  0.4em\relax MIT Press, 1999.

\bibitem{ytl}
ytopt, ``Autotuning applications at scale,''
  https://github.com/ytopt-team/ytopt-libensemble,
  https://github.com/ytopt-team/ytopt, 2023.

\bibitem{CV95}
\BIBentryALTinterwordspacing
C.~Cortes and V.~Vapnik, ``Support-vector networks,'' \emph{Machine Learning},
  vol.~20, no.~3, pp. 273--297, 1995. [Online]. Available:
  \url{https://doi.org/10.1007/BF00994018}
\BIBentrySTDinterwordspacing

\bibitem{WU23}
X.~Wu, P.~Balaprakash, M.~Kruse, J.~Koo, B.~Videau, P.~Hovland, V.~Taylor,
  B.~Gelts, S.~Jana, and M.~Hall, ``ytopt: Autotuning scientific applications
  for energy efficiency at large scales,'' in \emph{Proceedings of Cray User
  Group Conference 2023}, ser. CUG'23, Helsinki, Finland, May 7-11, 2023, 2023.

\bibitem{WU22}
\BIBentryALTinterwordspacing
X.~Wu, M.~Kruse, P.~Balaprakash, H.~Finkel, P.~Hovland, V.~Taylor, and M.~Hall,
  ``{Autotuning PolyBench benchmarks with LLVM Clang/Polly} loop optimization
  pragmas using {Bayesian} optimization,'' \emph{Concurrency and Computation:
  Practice and Experience}, vol. Volume 34, Issue 20, e6683, 2022. [Online].
  Available: \url{https://doi.org/10.1002/cpe.6683}
\BIBentrySTDinterwordspacing

\bibitem{CFS}
\BIBentryALTinterwordspacing
M.~Lindauer, K.~Eggensperger, M.~Feurer, A.~Biedenkapp, J.~Marben, P.~Müller,
  and F.~Hutter, ``Boah: A tool suite for multi-fidelity bayesian optimization
  \& analysis of hyperparameters,'' \emph{arXiv:1908.06756 {[cs.LG]}}, 2019.
  [Online]. Available: \url{https://automl.github.io/ConfigSpace/master/}
\BIBentrySTDinterwordspacing

\bibitem{WU24}
X.~Wu, J.~R. Tramm, J.~Larson, J.-L. Navarro, P.~Balaprakash, B.~Videau,
  M.~Kruse, P.~Hovland, V.~Taylor, and M.~Hall, ``Integrating ytopt and
  libensemble to autotune openmc,'' https://arxiv.org/pdf/2402.09222, 2024,
  arXiv:2402.09222 [cs.PF].

\bibitem{WU23a}
\BIBentryALTinterwordspacing
X.~Wu, P.~Paramasivam, and V.~Taylor, ``Autotuning apache tvm-based scientific
  applications using bayesian optimization,'' in \emph{Proceedings of the SC
  '23 Workshops of The International Conference on High Performance Computing,
  Network, Storage, and Analysis}, ser. SC-W '23.\hskip 1em plus 0.5em minus
  0.4em\relax New York, NY, USA: Association for Computing Machinery, 2023, p.
  29–35. [Online]. Available: \url{https://doi.org/10.1145/3624062.3626079}
\BIBentrySTDinterwordspacing

\bibitem{YAN23}
J.~Ma, ``Mixed-kernel heterojunction svm simulation,''
  https://github.com/JennyMa0517/mixed-kernel-heterojunction, 2023.

\bibitem{YSVM}
X.~Wu, ``Using ytopt and ytopt-libe to autotuning mixed kernel svm
  simulations,''
  https://github.com/ytopt-team/ytopt-libensemble/tree/main/ytopt-libe-svms,
  2024.

\bibitem{AK14}
J.~Ansel, S.~Kamil, K.~Veeramachaneni, J.~Ragan-Kelley, J.~Bosboom, U.-M.
  O'Reilly, and S.~Amarasinghe, ``{OpenTuner}: An extensible framework for
  program autotuning,'' in \emph{Proceedings of the 23rd International
  Conference on Parallel Architectures and Compilation Techniques}, ser.
  PACT'14.\hskip 1em plus 0.5em minus 0.4em\relax ACM, Aug. 2014, pp. 303--316.

\bibitem{LS21}
Y.~Liu, W.~M. Sid-Lakhdar, O.~Marques, X.~Zhu, C.~Meng, J.~W. Demmel, and X.~S.
  Li, ``{GPTune}: multitask learning for autotuning exascale applications,'' in
  \emph{Proceedings of PPoPP '21: Proceedings of the 26th ACM SIGPLAN Symposium
  on Principles and Practice of Parallel Programming}, ser. PPoPP'21.\hskip 1em
  plus 0.5em minus 0.4em\relax New York, NY, USA: Association for Computing
  Machinery, February 2021, pp. 234--246.

\bibitem{RT21}
R.~B. Roy, T.~Patel, V.~Gadepally, and D.~Tiwari, ``Bliss: Auto-tuning complex
  applications using a pool of diverse lightweight learning models,'' in
  \emph{Proceedings of the 42nd ACM SIGPLAN International Conference on
  Programming Language Design and Implementation}, ser. PLDI'21.\hskip 1em plus
  0.5em minus 0.4em\relax New York, NY, USA: Association for Computing
  Machinery, June 20--25, 2021, pp. 1280--1295.

\bibitem{TWORK}
V.~Taylor, ``Threadwork: A transformative co-design approach to materials and
  computer architecture research,'' https://www.anl.gov/threadwork, 2024.

\end{thebibliography}

\newpage
The submitted manuscript has been created by UChicago Argonne, LLC, Operator of Argonne National Laboratory ("Argonne"). Argonne, a U.S. Department of Energy Office of Science laboratory, is operated under Contract No. DE-AC02-06CH11357. The U.S. Government retains for itself, and others acting on its behalf, a paid-up nonexclusive, irrevocable worldwide license in said article to reproduce, prepare derivative works, distribute copies to the public, and perform publicly and display publicly, by or on behalf of the Government. The Department of Energy will provide public access to these results of federally sponsored research in accordance with the DOE Public Access Plan (http://energy.gov/downloads/doe-public-access-plan).

\end{document}